\documentclass[runningheads]{llncs}
\usepackage{graphicx}
\usepackage{svg}
\usepackage{adjustbox}
\usepackage{siunitx} 
\usepackage{color, soul}
\usepackage{amsfonts}
\usepackage{amsmath}
\usepackage{algorithm}
\usepackage{algorithmic}
\usepackage{booktabs}
\usepackage{multirow}
\usepackage{tabularx}
\newcolumntype{Y}{>{\centering\arraybackslash}X} 
\usepackage{amssymb}
\usepackage{pifont}
\usepackage{appendix}
\usepackage[normalem]{ulem}
\usepackage[colorlinks=true, linkcolor=black]{hyperref}
\usepackage{footmisc}
\begin{document}
\title{DLAFormer: An End-to-End Transformer For Document Layout Analysis}
\titlerunning{DLAFormer: An End-to-End Transformer For Document Layout Analysis}
%
%
%
%
\newcommand*\samethanks[1][\value{footnote}]{\footnotemark[#1]}
\author{
    Jiawei Wang\inst{1,2,}\thanks{Work done when Jiawei Wang and Kai Hu were interns at Multi-Modal Interaction Group, Microsoft Research Asia, Beijing, China.}$^,$\thanks{Equal contribution. Correspondence to \email{wangjiawei@mail.ustc.edu.cn}.} \and
    Kai Hu\inst{1,2,}\samethanks[1]$^,$\samethanks[2] \and
    Qiang Huo\inst{2}
}
\authorrunning{Wang et al.}
\institute{
Department of EEIS, University of Science and Technology of China, Hefei, China \and
Microsoft Research Asia, Beijing, China \\
\email{wangjiawei@mail.ustc.edu.cn, hk970213@mail.ustc.edu.cn, qianghuo@microsoft.com}}




\def \Ours {DLAFormer}

\maketitle              
\begin{abstract}
Document layout analysis (DLA) is crucial for understanding the physical layout and logical structure of documents, serving information retrieval, document summarization, knowledge extraction, etc. However, previous studies have typically used separate models to address individual sub-tasks within DLA, including table/figure detection, text region detection, logical role classification, and reading order prediction. In this work, we propose an end-to-end transformer-based approach for document layout analysis, called \Ours, which integrates all these sub-tasks into a single model. To achieve this, we treat various DLA sub-tasks (such as text region detection, logical role classification, and reading order prediction) as relation prediction problems and consolidate these relation prediction labels into a unified label space, allowing a unified relation prediction module to handle multiple tasks concurrently. Additionally, we introduce a novel set of {\em type-wise queries} to enhance the physical meaning of content queries in DETR. Moreover, we adopt a coarse-to-fine strategy to accurately identify graphical page objects. Experimental results demonstrate that our proposed \Ours\ outperforms previous approaches that employ multi-branch or multi-stage architectures for multiple tasks on two document layout analysis benchmarks, DocLayNet and Comp-HRDoc.

\keywords{Document Layout Analysis \and Relation Prediction \and Unified Label Space}
\end{abstract}
\section{Introduction}

Document layout analysis (DLA) is the process of recovering document physical and logical structures from document images, including physical layout analysis and logical structure analysis \cite{doermann2014handbook}. Given input document images, physical layout analysis aims at identifying physical homogeneous regions of interest (also called page objects), such as graphical page objects like tables, figures, mathematical formulas, and different types of text regions. Logical structure analysis aims at assigning a logical role to each identified region (e.g., title, section heading, header, footer, paragraph) and determining their logical relationships (e.g., reading order relationships) \cite{zhong2023hybrid}. Document layout analysis plays a crucial role in advancing document understanding, enabling various applications like document digitization, conversion, archiving, and retrieval. However, the persistent challenge stems from the diverse content and intricate complexities inherent in document layouts, rendering it a formidable problem.

Document layout analysis encompasses numerous intricate sub-tasks, such as graphical page object detection, text region detection, logical role classification, reading order prediction, etc. While prior research \cite{zhong2019publaynet,pfitzmann2022doclaynet,li2020readingorder,hu2023kvpformer,wang2023dqdetr,HU2024mathdetection} has predominantly concentrated on these individual sub-tasks by employing specialized models, there is a growing interest in the consolidation of these subtasks. The integration of DLA sub-tasks into a cohesive framework stands as a promising avenue for enhancing both efficiency and effectiveness. In line with this vision, recent initiatives have begun to adopt a more holistic approach, designing versatile multi-branch or multi-stage frameworks capable of addressing several sub-tasks concurrently. DocParer \cite{rausch2021docparser} is a two-stage end-to-end system for parsing document renderings into hierarchical document structures, which first employs Mask R-CNN \cite{he2017mask} to detect all document entities and devises a set of rules to predict relationships between them. Built upon it, DSG \cite{rausch2023dsg} replaced the rule-based relation prediction module with an LSTM-based relation prediction network to make the whole system end-to-end trainable. Zhong et al. \cite{zhong2023hybrid} introduced a two-branch hybrid approach for document layout analysis, which combined the advantages of top-down and bottom-up based methods and addressed several sub-tasks such as graphical page object detection, text region detection and reading order prediction within text regions concurrently. To analyze document structures, Ma et al. \cite{Ma2023HRDoc} proposed a new dataset, HRDoc, for a novel task: Hierarchical Document Structure Reconstruction. They employed a two-stage approach to classify the logical role of each text unit and predict the hierarchical relationships between these units. Building on HRDoc, Detect-Order-Construct \cite{wang2023doc} established a comprehensive benchmark named Comp-HRDoc, encompassing page object detection, reading order prediction, table of contents extraction, and hierarchical structure reconstruction concurrently. A three-stage framework based on tree construction is proposed to handle these tasks. While these methods exhibit promise on certain DLA benchmarks, multi-branch or multi-stage frameworks may introduce cascading errors when addressing these DLA sub-tasks sequentially. Such approaches also pose scalability challenges and struggle to accommodate additional tasks.

In this paper, we present \Ours, an end-to-end transformer-based approach for document layout analysis. In contrast to conventional approaches that typically employ multi-branch or multi-stage architectures, \Ours{} simplifies the training process by casting various DLA sub-tasks (such as text region detection, logical role classification, and reading order prediction) as relation prediction problems. The labels of these tasks are unified into a single, comprehensive label space, thereby streamlining the learning mechanism. \Ours{} is constructed upon an encoder-decoder Transformer architecture inspired by DETR \cite{carion2020end} and is capable of inferring all relationships concurrently, enabling an end-to-end training process through a unified relation prediction module. Inspired by Deformable DETR \cite{zhu2020deformable}, we introduce novel \emph{type-wise queries} to capture categorical information of diverse page objects. This enhances the physical meaning of content queries in transformer decoder, refining our model's capabilities for these DLA sub-tasks. Additionally, \Ours{} adopts a coarse-to-fine strategy \cite{zhu2020deformable} to precisely identify graphical page objects within document images. \Ours{} offers several advantages. 
Firstly, it demonstrates outstanding scalability, effortlessly enabling the seamless integration of new DLA tasks into the framework through its flexible query mechanism and the unified label space approach. The incorporation of new relationship labels can seamlessly merge into the unified label space, and novel types of elements can be seamlessly treated as new types of queries. 
Secondly, the unified label space approach empowers a unified relation prediction module to predict all relationships in a single pass. This proves more effective and efficient in capturing potential relationships among these layout units. Thirdly, \Ours{} enhances units by incorporating robust contextual information through self-attention mechanisms and enabling attentive consideration of global and local document layout information via cross-attention mechanisms. Experimental results show that our \Ours{} surpasses prior approaches using multi-branch or multi-stage architectures for various tasks on two document layout analysis benchmarks, DocLayNet \cite{pfitzmann2022doclaynet} and Comp-HRDoc \cite{wang2023doc}.

The main contributions of this paper can be summarized as follows:

\begin{itemize}

\item We introduce \Ours, a novel end-to-end transformer-based approach for Document Layout Analysis, consolidating tasks including graphical page object detection, text region detection, logical role classification, and reading order prediction within a unified model.

\item We treat various document layout analysis sub-tasks (such as text region detection, logical role classification, and reading order prediction) as relation prediction problems and propose a unified label space approach to enabling a unified relation prediction module to effectively and efficiently handle these tasks concurrently.

\item Experimental results demonstrate that \Ours{} outperforms previous approaches that employ multi-branch or multi-stage architectures for multiple tasks on two document layout analysis benchmarks.

\end{itemize}

\section{Related Work}

\subsection{Page Object Detection}

Page object detection (POD) \cite{gao2017pod} plays a pivotal role in document layout analysis. It encompasses the identification and classification of logical objects, such as tables, figures, formulas, and paragraphs, within document pages. Deep learning-based POD approaches can be broadly classified into two categories: top-down based methods, and bottom-up based methods.

\textbf{Top-down based methods} leverage the latest top-down object detection or instance segmentation frameworks to address the page object detection problem. Yi et al. \cite{yi2017cnn} and Oliveira et al. \cite{augusto2017fast} paved the way by adapting R-CNN \cite{girshick2014rich} to document images, but performance stagnated due to limitations in traditional region proposal generation. Subsequent research \cite{zhong2019publaynet,saha2019graphical,li2022dit,biswas2022docsegtr,pfitzmann2022doclaynet,yang2022transformer} has embraced more sophisticated object detectors like Faster R-CNN \cite{ren2015faster}, Mask R-CNN \cite{he2017mask}, Cascade R-CNN \cite{cai2019cascade}, SOLOv2 \cite{wang2020solov2}, YOLOv5 \cite{yolov5}, and Deformable DETR \cite{zhu2020deformable}.
Enhancing detector performance has fueled innovative techniques. Zhang et al. \cite{zhang2021vsr} fused visual features with multimodal text embeddings and integrated a GNN-based relation module into a Faster/Mask R-CNN model, addressing candidate interactions. Shi et al. \cite{shi2022lateral} proposed a novel backbone network for improved feature extraction, while Yang et al. \cite{yang2022transformer} utilized a robust Swin Transformer \cite{liu2021swin} to boost Mask R-CNN and Deformable DETR performance. Recent advancements \cite{gu2022unified,li2022dit,huang2022layoutlmv3,maity2023selfdocseg} employed transformers pre-trained via self-supervised learning on vast document image datasets, integrating with top-down based object detection models like Mask R-CNN or Cascade R-CNN to effectively identify page objects. These efforts have achieved state-of-the-art benchmark performance, though challenges remain, particularly in detecting small-scale text regions.

\textbf{Bottom-up based methods} (e.g., \cite{li2018page,li2020page,wang2022post,liu2022unified,long2022towards}) typically represent each document page as a graph, where its nodes correspond to primitive page objects (e.g., words, text-lines, connected components), and its edges denote relationships between neighboring primitive page objects. The detection of page objects is then formulated as a graph labeling problem. For instance, Li et al. \cite{li2018page} employed image processing techniques to initially generate line regions, followed by the application of two CRF models to classify these regions into distinct types and predict whether pairs of line regions belong to the same instance, based on visual features extracted by CNNs. More recently, Wang et al. \cite{wang2023graphical} framed the page object detection problem as a graph segmentation and classification problem and introduced a lightweight graph neural network. While these bottom-up based methods can effectively detect small-scale text regions, accurately locating the graphical page objects based on text units within graphical objects remains challenging.

While achieving remarkable results on several benchmark datasets, both top-down and bottom-up based approaches exhibit inherent limitations. In light of these constraints, Zhong et al. \cite{zhong2023hybrid} introduced a novel two-branch hybrid approach that merges the strengths of both methods. In contrast to the conventional use of multi-branch or multi-stage structures for document layout analysis, our proposed \Ours{} integrates multiple sub-tasks into a unified end-to-end model. This not only combines the advantages of the hybrid approach but also demonstrates robust scalability and performance.

\subsection{Reading Order Prediction}

The objective of reading order prediction is to determine the appropriate reading sequence for documents. Generally, humans tend to read documents in a left-to-right and top-to-bottom manner. However, such simplistic sorting rules may prove inadequate when applied to complex documents with tokens extracted by OCR tools. 

Early approaches to reading order prediction are mainly based on heuristic sorting rules \cite{breuel2003high,Meunier2005optimized,FerilliP2015abstract}. Despite their effectiveness in certain scenarios, these rule-based methods can be prone to failure when confronted with out-of-domain cases. In recent years, deep learning models have emerged for reading order prediction. Li et al. \cite{li2020readingorder} proposed an end-to-end OCR text reorganizing model, using a graph convolutional encoder and a pointer network decoder to reorder text blocks. LayoutReader \cite{wang2021layoutreader} introduced a benchmark dataset called ReadingBank, which contains reading order, text, and layout information, and employed a transformer-based architecture on spatial-text features to predict the reading order sequence of words. However, the decoding speed of these auto-regressive-based methods is limited when applied to rich text documents. Recently, Quir{'{o}}s et al. \cite{Quiros2022reading} followed the idea of assuming a pairwise partial order at the element level from \cite{breuel2003high} and proposed two new reading-order decoding algorithms for reading order prediction on handwritten documents. Detect-Order-Construct \cite{wang2023doc} incorporated a similar relation prediction method akin to the one introduced in \cite{zhong2023hybrid} for reading order prediction, exhibiting much better performance in their benchmark datasets. In our work, we introduced a novel approach employing a unified label space to simultaneously address diverse relation prediction tasks.

\subsection{DETR and Its Variants}

DETR \cite{carion2020end} is a novel transformer-based \cite{transformer2017} object detection algorithm, which introduced the concept of object query and set prediction loss to object detection. These novel attributes make DETR get rid of many manually designed components in previous CNN-based object detectors like anchor design and non-maximum suppression (NMS). However, DETR has its own issues: 1) Slow training convergence; 2) Unclear physical meaning of object queries; 3) Hard to leverage high-resolution feature maps due to high computational complexity. Deformable DETR \cite{zhu2020deformable} proposed several effective techniques to address these issues, including learnable reference points, deformable attention module, iterative bounding box refinement strategy, and a two-stage DETR framework. Inspired by the concept of reference point in Deformable DETR, some follow-up works \cite{meng2021conditional,wang2022anchor,liu2022dab,gao2021fast} attempted to address the slow convergence issue by giving spatial priors to the object query. DN-DETR \cite{li2022dn} recognized the bipartite matching algorithm in Hungarian loss as a factor causing slow convergence and proposed a denoising-based training method to expedite DETR convergence. Subsequently, DINO \cite{zhang2022dino} further improved the denoising-based training method and introduced a mixed query selection strategy to enhance overall performance on the object detection task. In this paper, we adopt Deformable DETR as the primary architecture of \Ours{} and introduce novel type-wise queries to enhance the physical meaning of content queries in the transformer decoder.
\section{Problem Definition}
\label{sec:problem}

\begin{figure}[t]
    \centering
    \includegraphics[width=1.0\linewidth]{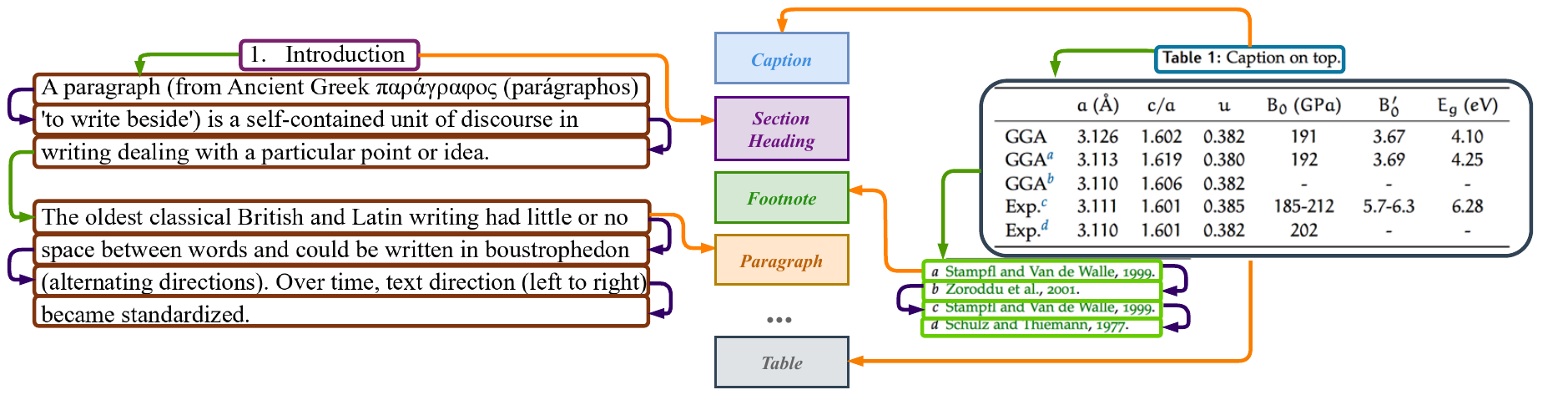}
    \caption{An example of our problem definition for document layout analysis. Purple arrow: \textit{intra-region relationship}; green arrow: \textit{inter-region relationship}; orange arrow: \textit{logical role relationship}. Best viewed in color.}
    
    \label{fig:problem}
\end{figure}

A document image inherently comprises diverse regions, encompassing both Text Regions and Non-Text Regions. A Text Region serves as a semantic unit of written content, comprising text-lines arranged in a natural reading sequence and associated with a logical label, such as paragraph, list/list item, title, section heading, header, footer, footnote, and caption. Non-text regions typically include graphical elements like tables, figures, and mathematical formulas. Multiple logical relationships often exist between these regions, with the most common being the reading order relationship.

Consequently, we define three distinct types of relationships: \textit{intra-region relationship}, \textit{inter-region relationship}, and \textit{logical role relationship}. These relationships aim to group basic text units, such as text-lines, into coherent text regions and explore the logical connections between these regions.
Specifically, given a document image $D$ composed of $N$ text-lines $[T_1, T_2, ..., T_N]$ and $M$ graphical objects $[G_1, G_2, ..., G_M]$, we define the relationships as follows:
\begin{itemize}
    \item As illustrated in Fig.~\ref{fig:problem}, consider each text region, composed of several text-lines arranged in a natural reading sequence. We establish \textit{intra-region relationships} for all adjacent text-lines within the same text region. For text regions consisting of a single text line, we designate the relationship of this text line as self-referential.
    \item To delve into the logical connections among these text regions and non-text regions, we construct \textit{inter-region relationships} between all pairs of regions that exhibit logical connections. For instance, as depicted in Fig.~\ref{fig:problem}, we establish an \textit{inter-region relationship} between two adjacent paragraphs and between a table and its corresponding caption or footnote.
    \item As shown in Fig.~\ref{fig:problem}, we delineate various logical role units, including caption, section heading, paragraph, title, etc. Given that each text region is assigned a logical role, we establish a \textit{logical role relationship} between each text line in the text region and its corresponding logical role unit.
\end{itemize}

By defining these relationships, we frame various DLA sub-tasks (such as text region detection, logical role classification, and reading order prediction) as relation prediction challenges and merge the labels of different relation prediction tasks into a unified label space, thereby employing a unified model to handle these tasks concurrently. 

\section{Methodology}

\begin{figure}[t]
    \centering
    \includegraphics[width=1.0\linewidth]{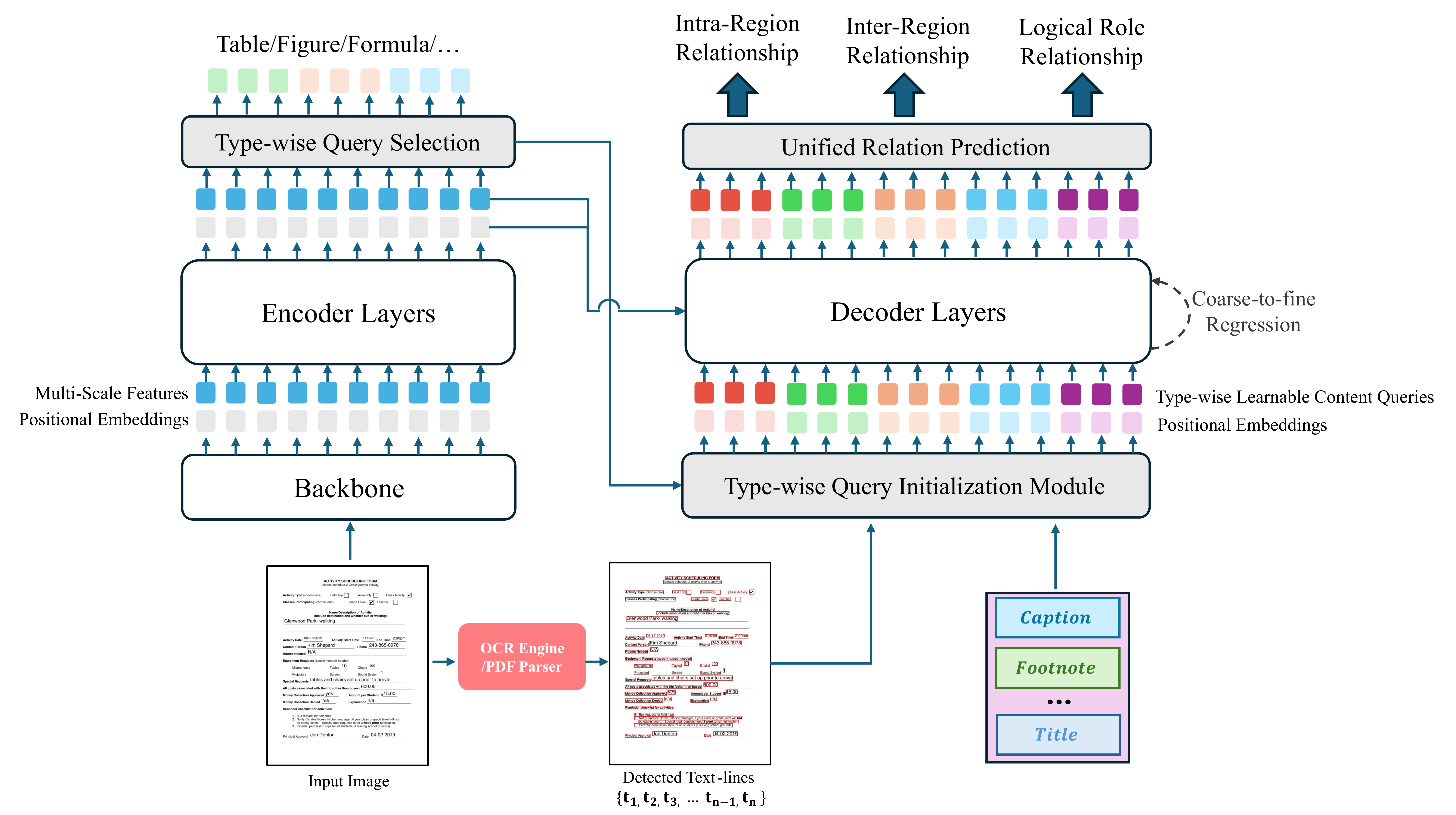}
    \caption{Overall architecture of our \Ours{} for document layout analysis.}
    
    \label{fig:framework}
\end{figure}


\subsection{Model Overview}

Leveraging the new perspective of our proposed unified label space, we propose \Ours{}, an end-to-end transformer-based approach for document layout analysis. Following the DETR-like model architecture, \Ours{} incorporates a backbone, a multi-layer Transformer encoder, a multi-layer Transformer decoder, a unified relation prediction head, and a coarse-to-fine detection head. The overall pipeline is depicted in Fig.~\ref{fig:framework}. Given a document image, we start by extracting multi-scale features $\{C_3,C_4,C_5\}$ using backbones such as ResNet \cite{he2016deep} or Swin Transformer \cite{liu2021swin}. These features are then fed into the Transformer encoder, along with corresponding positional embeddings. To enhance computational efficiency in handling multi-scale features, we integrate a deformable transformer encoder to enhance these extracted features. Following the feature enhancement in the encoder, we employ a \emph{type-wise query selection strategy} to acquire a reference box and a category label for each potential graphical object proposal, which will be described in Sec.~\ref{sec:CQS}. For text-lines within the given document image, we utilize a PDF parser or OCR engine to extract their bounding boxes. These graphical object proposals and text-lines will be served as queries and input into the Transformer decoder. To bolster the physical meaning of these queries and facilitate adaptive feature capture from distinct regions for various types of queries, we introduce a \emph{type-wise query initialization module} to initialize type-wise queries as content queries for the subsequent decoder, as elaborated in Sec.~\ref{sec:TLQM}. Following the acquisition of positional queries, content queries, and their reference boxes, we leverage a deformable transformer decoder to enhance these queries. This involves incorporating a self-attention module to model interactions among these queries, while a deformable cross-attention module is employed to capture both global and local layout information from multi-scale feature maps. Taking inspiration from Deformable DETR \cite{zhu2020deformable}, we adopt a coarse-to-fine regression strategy to iteratively refine the reference boxes of graphical object queries layer-by-layer. Finally, to unravel the logical connections between these queries, we introduce a \emph{unified relation prediction head} that effectively and efficiently handles relation prediction tasks concurrently. Further details on this aspect will be provided in Sec.~\ref{sec:URPH}.

\subsection{Type-wise Query Selection}
\label{sec:CQS}

In DETR \cite{carion2020end} and DAB-DETR \cite{liu2022dab}, the decoder queries are static embeddings that do not incorporate any encoder features specific to an individual image. These models learn anchors or positional queries directly from the training data and initialize content queries as all-zero vectors. To enhance prior knowledge for improved decoder queries, Deformable DETR \cite{zhu2020deformable} introduces a mechanism where multi-scale features from the encoder are fed into an auxiliary detection head with a binary classifier. The top-$K$ features are then selected based on the objectiveness (class) score of each encoder feature to initialize both positional and content queries. Simultaneously, the corresponding predicted boxes are employed to initialize reference boxes. To address potential ambiguities and confusion for the decoder arising from selected encoder features, DINO \cite{zhang2022dino} proposes a mixed query selection approach. This approach selectively enhances only the positional queries with the top-$K$ selected features, while keeping the content queries learnable as before.

Although the approach in DINO has brought significant improvements, the unclear physical meaning of the learnable content query remains an issue. To address this, we introduce a type-wise query selection strategy, which involves leveraging potential class information to initialize the content query, thereby departing from the use of ``static" content queries. Given the substantial differences in visual features among various types of graphical objects, such as formulas, tables, and figures, initializing content queries with category information will enable these queries to adaptively capture crucial features in the decoder.
Specifically, we substitute the binary classifier in the auxiliary detection head with a multi-class classifier to discern the class of each selected feature. While the predicted reference boxes are still utilized for initializing the positional queries, the predicted category is directed toward the subsequent type-wise query initialization module. Within this module, a type-wise learnable content query is assigned for each type of query.

\subsection{Type-wise Query Initialization Module}
\label{sec:TLQM}

We introduce a type-wise query initialization module to standardize the modeling of logical relationships among different queries, ensuring a uniform input into the decoder. As depicted in Fig.~\ref{fig:framework}, the type-wise query initialization module takes three components as input: reference boxes and categories of graphical object proposals, bounding boxes of extracted text-lines, and pre-defined logical role types. For graphical object proposals from the encoder, we initiate the positional queries by applying sine positional encoding \cite{transformer2017} to the reference boxes. Simultaneously, we define learnable features for each category and initialize content queries by selecting the corresponding features based on the category. Similarly, for text lines, we adopt a comparable approach. We begin by initializing positional queries based on the bounding box. Then, we define a distinct learnable feature for these text lines, which serves as the initialization for the content queries. Previous approaches to logical role classification typically used a static parameter classifier, treating it as a straightforward multi-class classification task. Inspired by dynamic algorithms \cite{jia2016dynamic,tian2020conditional}, we reformulate logical role classification as a relationship prediction problem. In this framework, we establish both positional and content queries for predefined logical roles, such as titles, section headings, captions, etc. This approach allows the logical role query to dynamically adapt its feature extraction process to the specifics of each image. Each basic unit within an image is then tasked with predicting a pointer toward these dynamic logical role queries, enhancing the model's adaptability and responsiveness to unique image content. 
Additionally, we use learnable features for each logical role type as content query initialization. For uniformity in query structure, corresponding learnable reference boxes are defined for each logical role type. During training, we introduce auxiliary supervision by utilizing the union boxes of all queries belonging to a specific logical role as the reference box supervision for that role.

\subsection{Unified Relation Prediction Head}
\label{sec:URPH}
\begin{figure}[t]
    \centering
    \includegraphics[width=1.0\linewidth]{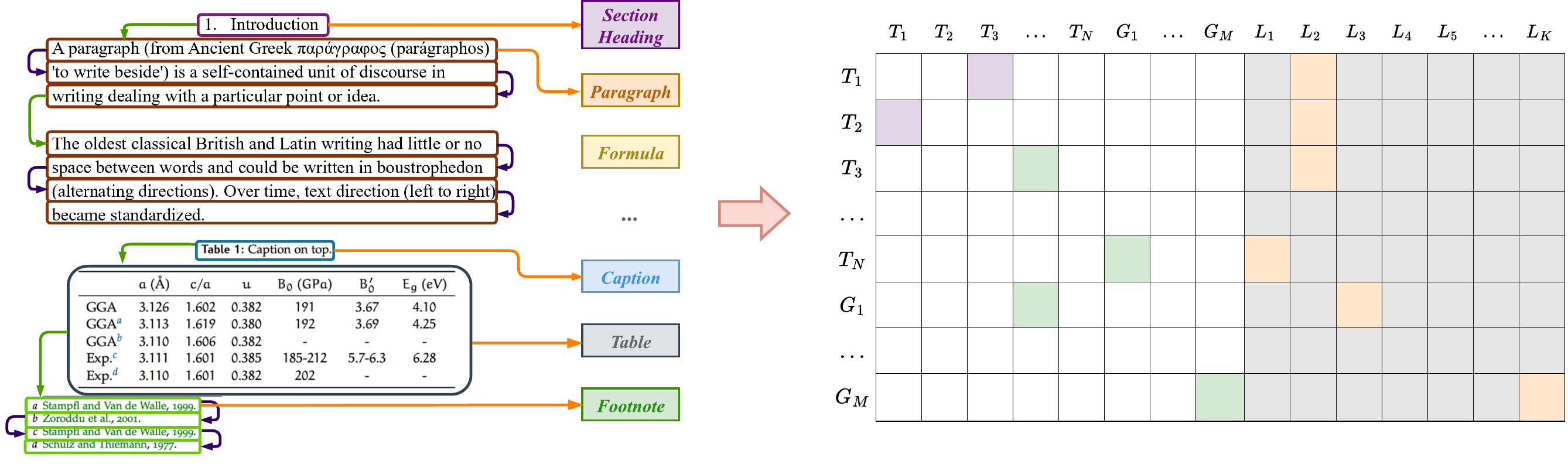}
    \caption{The unified label space in  \Ours{}. $T_i$ denotes \emph{Text-line queries}, $G_i$ denotes \emph{Graphical object queries}, and $L_i$ denotes \emph{Logical role queries}. The purple grids illustrate \emph{intra-region relationships}, the green grids represent \emph{inter-region relationships}, and the orange grids signify \emph{logical role relationships}.}
    \label{fig:unified}
\end{figure}

Following the enhancement process in the decoder layers, three types of queries including text-line queries, graphical object queries, and logical role queries, are input into our unified relation prediction head to uncover the logical connections between these queries. As discussed in Sec.~\ref{sec:problem}, we define three distinct types of relationships: \emph{intra-region relationships}, \emph{inter-region relationships}, and \emph{logical role relationships}. To effectively and efficiently handle these relation prediction tasks concurrently, we introduce a unified label space approach, as illustrated in Fig.~\ref{fig:unified}. Specifically, we define a label matrix $M \in \mathbb{Z}^{H\times W}$, where each element in the $i$-th row and $j$-th column can take on four possible values. Taking Fig.~\ref{fig:unified} as an example, the empty cell in the label matrix signifies no logical relationship pointing from the $i$-th query to the $j$-th query. The other three types of cells represent three pre-defined relationships, each with its distinct interpretation. With this unified label space, our unified relation prediction head consists of two modules: a \emph{relation prediction module} and a \emph{relation classification module}.

\subsubsection{Relation Prediction Module.} Taking into account the logical relationships between text-line/graphical object queries and their connections to logical role queries, we group all text-line and graphical object queries as $\{q_1, q_2, ..., q_H\}$ and logical role queries as $\{q_{H+1}, q_{H+2}, ..., q_W\}$. We calculate the scores $s_{ij}$, representing the probability of a logical relationship from $q_i$ to $q_j$, as follows:
\begin{gather}
\label{relation_score}
f_{ij} = FC^r_q(q_i) \circ FC^r_k(q_j), i\leq H, j \leq W \\
s_{ij} = \begin{cases}
\frac{\exp(f_{ij})}{\sum_{j=1}^{H} \exp(f_{ij})}, & j\leq H \\
\\
\frac{\exp(f_{ij})}{\sum_{j=H+1}^{W} \exp(f_{ij})},& H < j \leq W \\
\end{cases}
\end{gather}
where each of $FC^r_q$ and $FC^r_k$ represents a single fully-connected layer with 1,024 nodes, serving to map $q_i$ and $q_j$ into distinct feature spaces; $\circ$ denotes dot product operation. 

\subsubsection{Relation Classification Module.} We use a multi-class classifier to determine the relation type between $q_i$ and $q_j$ by computing the probability distribution across various classes. The process unfolds as follows:
\begin{gather}
p_{ij} = BiLinear(FC^c_q(q_{i}), FC^c_k(q_{j})), i\leq H, j \leq W \\
c_{ij} = argmax(p_{ij})
\end{gather}
where both $FC^c_q$ and $FC^c_k$ represent single fully-connected layers with 1,024 nodes; $BiLinear$ signifies the bilinear classifier; and $argmax$ identifies the index $c_{ij}$ with the highest value in the probability distribution $p_{ij}$, serving as the predicted relation type.

\subsection{Optimization}

In \Ours{}, we utilize the same detection heads as in Deformable DETR \cite{zhu2020deformable}, except for replacing the binary classifier with a multi-class classifier in the encoder. The optimization of these detection heads follows the training paradigm outlined in Deformable DETR. In training, we incorporate a shared unified relation prediction head at each layer of the decoder to facilitate training. Inspired by \cite{zhong2023hybrid}, all of these relation prediction modules and relation classification modules consistently utilize softmax cross-entropy as their loss function for effective optimization. The overall loss of our model is determined by aggregating the individual losses from each prediction head.
\section{Experiments}

\subsection{Datasets and Evaluation Protocols}

We conduct experiments on two document layout analysis benchmarks, Comp-HRDoc \cite{wang2023doc} and DocLayNet \cite{pfitzmann2022doclaynet}, to validate the effectiveness of \Ours{}.

\textbf{Comp-HRDoc} \cite{wang2023doc} is an extensive benchmark, specifically designed for comprehensive hierarchical document structure analysis. It encompasses a range of document layout analysis tasks, including page object detection, reading order prediction, table of contents extraction, and hierarchical structure reconstruction. Comp-HRDoc is constructed based on the HRDoc-Hard dataset \cite{Ma2023HRDoc}, which includes 1,000 documents for training and 500 documents for testing. Given that our model is presently designed to integrate various page-level document layout analysis tasks into a unified framework, our evaluation centers on two specific tasks: page object detection and reading order prediction. In the page object detection task, covering sub-tasks like graphical page object detection, text region detection, and logical role classification, we use COCO-style Segmentation-based mean Average Precision (Segm. mAP) as the evaluation metric. Additionally, the assessment of the reading order prediction task includes a Reading Edit Distance Score (REDS), which encompasses Text Region REDS and Graphical Region REDS as proposed in \cite{wang2023doc}.

\textbf{DocLayNet} \cite{pfitzmann2022doclaynet} stands out as a challenging human-annotated dataset for page object detection recently introduced by IBM. This dataset comprises 69,375 pages for training, 6,489 for testing, and 4,999 for validation. Spanning various document categories such as Financial reports, Patents, Manuals, Laws, Tenders, and Scientific Papers, DocLayNet includes 11 pre-defined types of page objects. These objects encompass Caption, Footnote, Formula, List-item, Page-footer, Page-header, Picture, Section-header, Table, Text (i.e., Paragraph), and Title. The evaluation metric of DocLayNet is the COCO-style box-based mean Average Precision (mAP).

In addition to document images, both datasets provide OCR files containing information about bounding boxes and the reading order of text-lines. Notably, Comp-HRDoc has pre-filtered text-lines within graphical objects. However, OCR files of DocLayNet present a challenge with a considerable number of text-lines within graphical objects, creating a potential long-tail problem for unified models in logical role classification and relation prediction.

\subsection{Implementation Details}

Our approach is implemented using PyTorch v1.11, and all experiments are conducted on a workstation equipped with 16 Nvidia Tesla V100 GPUs (32 GB memory). In \Ours{}, both the transformer encoder and decoder are configured with 3 layers. Both are designed with the number of heads, the dimension of the hidden state, and the dimension of the feedforward network set as 8, 256, and 1024, respectively. For the initialization of graphical object queries, we opt for the top-$50$ encoder features on Comp-HRDoc and the top-$100$ features on DocLayNet. In training phase, the parameters of CNN backbone network are initialized with a ResNet-50 model \cite{he2016deep} pre-trained on ImageNet classification task. The optimization process employs AdamW algorithm \cite{loshchilov2017decoupled} with a mini-batch size of 4, trained for 40 epochs on Comp-HRDoc and 24 epochs on DocLayNet. For the CNN backbone network, the learning rate and weight decay are set to 1e-5 and 1e-4, respectively, while for the transformer architecture, they are set to 1e-4 and 1e-4. Other hyper-parameters of AdamW, including betas and epsilon, are set as (0.9, 0.999) and 1e-8. Additionally, a multi-scale training strategy is adopted, where the shorter side of each image is randomly rescaled to a length chosen from [320, 416, 512, 608, 704, 800], while the longer side should not exceed 1024. In testing phase, we set the shorter side of input image as 512. As mentioned earlier, the large number of text-lines within graphical objects in DocLayNet gives rise to a serious long-tail problem in classification during training. Therefore, drawing inspiration from \cite{menon2020long}, we opt for the logit-adjusted softmax cross-entropy loss when training on DocLayNet.

\subsection{Comparisons with Prior Arts}

In this section, we compare \Ours{} with several state-of-the-art methods on Comp-HRDoc and DocLayNet to showcase the effectiveness of \Ours{}.

\begin{table}[t]  
\centering  
\caption{Performance comparisons in page object detection task on Comp-HRDoc (in \%). The symbol $^\dag$ represents the results of our replication.}
\begin{adjustbox}{width=\textwidth}
\begin{tabular}{  
    l|cccccccccccc|c
}  
\toprule  
Method & {Title} & {Auth} & {Mail} & {Affil} & {Sect} & {Para} & {Table} & {Fig} & {Cap} & {Foot} & {Head} & {Footn} & {Segm. mAP} \\  
\midrule  
Mask2Former-R18 \cite{cheng2022mask2former} & 84.4 & 70.9 & 52.0 & 62.1 & 76.5 & 71.3 & 79.8 & 86.2 & 71.8 & 56.7 & 54.9 & 66.5 & 69.4 \\  
DOC-R18 \cite{wang2023doc} & 94.5 & 79.4 & 51.6 & 70.9 & 89.7 & 83.4 & \textbf{80.3} & \textbf{86.4} & 89.0 & 95.9 & 95.1 & 84.7 & 83.4 \\  
DLAFormer-R18 & \textbf{96.9} & \textbf{89.4} & \textbf{70.1} & \textbf{84.5} & \textbf{91.8} & \textbf{86.3 }& 79.5 & 83.9 & \textbf{92.1} & \textbf{97.4} & \textbf{98.3} & \textbf{90.8} & \textbf{88.4} \\  
\midrule
Mask2Former-R50 \cite{cheng2022mask2former} & 85.4 & 74.7 & 66.7 & 69.2 & 78.2 & 74.6 & \textbf{80.5} & \textbf{86.6} & 76.1 & 62.7 & 59.0 & 69.0 & 73.6 \\  
DOC-R50$^\dag$  \cite{wang2023doc} & 95.3 & 85.3 & 65.0 & 77.6 & 91.0 & 85.0 & 80.4 & \textbf{86.6} & 91.9 & 96.4 & 95.6 & 88.7 & 86.5 \\  
DLAFormer-R50 & \textbf{96.8} & \textbf{91.1} & \textbf{75.4} & \textbf{86.0} & \textbf{92.8} & \textbf{87.7} & 79.4 & 85.5 & \textbf{92.5} & \textbf{97.8} & \textbf{98.2} & \textbf{91.7} & \textbf{89.6} \\  
\bottomrule  
\end{tabular}  
\end{adjustbox}

\label{tab:comp-hrdoc-pod}
\end{table}

\begin{table}[t]
\centering
\caption{Performance comparisons in reading order prediction task on Comp-HRDoc (in \%). The symbol $^\dag$ represents the results of our replication.}
\fontsize{9}{11}\selectfont
\setlength{\tabcolsep}{10pt}
\begin{tabular}{lcc}
\toprule
{ Method}         & { \begin{tabular}[c]{@{}c@{}}Text Region \\ REDS\end{tabular}} & { \begin{tabular}[c]{@{}c@{}}Graphical Region \\ REDS\end{tabular}} \\ \midrule
{ Lorenzo et al. \cite{Quiros2022reading}} & { 77.4}                                                                              & { 85.8}                                                                                  \\ \midrule
{ DOC-R18 \cite{wang2023doc}}             & { 93.2}                                                                              & { 86.4}                                                                                  \\
{ DLAFormer-R18}       & { \textbf{96.3}}                                                                      & { \textbf{89.6}}                                                                          \\
\midrule
{ DOC-R50$^\dag$ \cite{wang2023doc}}             & { 94.4}                                                                              & { 88.6}                                                                                  \\
{ DLAFormer-R50}       & { \textbf{96.6}}                                                                      & { \textbf{90.0}}                                                                          \\ \bottomrule
\end{tabular}

\label{tab:comp-hrdoc-reading}
\end{table}

\subsubsection{Comp-HRDoc.} As our method is a purely vision-based framework, we conduct comparisons between \Ours{} and the vision-only Detector-Order-Construct (DOC) \cite{wang2023doc} in two document layout analysis tasks, page object detection and reading order prediction, on Comp-HRDoc dataset. As illustrated in Table~\ref{tab:comp-hrdoc-pod}, \Ours{} surpasses the top-down based approach, Mask2Former \cite{cheng2022mask2former}. While both DOC and \Ours{} employ a combination of top-down and bottom-up approaches, \Ours{} consistently outperforms in categories like Title, Author, Mail, Affiliate, and Paragraph. This highlights its ability to effectively model interactions among text units and capture global and local layout information from multi-scale feature maps through attention mechanisms within the transformer decoder.
Notably, DOC utilizes Mask2Former with feature maps $\{C_2, C_3, C_4, C_5\}$ for table and figure detection. In contrast, \Ours{} achieves comparable performance with fewer feature maps, specifically $\{C_3, C_4, C_5\}$, in identifying tables and figures. Additionally, DOC employs a two-stage approach for predicting reading order relationships, necessitating additional parameters. On the contrary, \Ours{} utilizes a unified label space to predict all relationships simultaneously, effectively reducing cascading errors and yielding superior results, as indicated in Table~\ref{tab:comp-hrdoc-reading}.

\begin{table}[t]

\fontsize{9}{11}\selectfont
\setlength{\tabcolsep}{4pt}

\caption{Performance comparisons on DocLayNet testing set (in \%). The results of Mask R-CNN, Faster R-CNN, and YOLOv5 are extracted from \cite{pfitzmann2022doclaynet}, while the results of DOC are sourced from \cite{wang2023doc}.}
\begin{center}
\begin{tabular}{l|c|ccccccc}
\hline \# & Human & \begin{tabular}[c]{@{}c@{}}Mask\\ R-CNN\end{tabular} & \begin{tabular}[c]{@{}c@{}}Faster\\ R-CNN\end{tabular} & YOLOv5 & DINO & DOC & \Ours \\
\hline Caption & $\operatorname{84-89}$ & $71.5$ & $70.1$ & $77.7$ & 85.5 &  83.2 & \textbf{89.7}\\
Footnote & $\operatorname{83-91}$ & $71.8$ & $73.7$ & $\textbf{77.2}$ & $69.2$ & 67.8 & 63.1 \\
Formula & $\operatorname{83-85}$ & $63.4$ & $63.5$ & $ 66.2 $ & $63.8$ & 63.9 & \textbf{81.1} \\
List-item & $\operatorname{87-88}$ & $80.8$ & $81.0$ & $86.2$ & $80.9$ & \textbf{86.9} & 86.0\\
Page-footer & $\operatorname{93-94}$ & $59.3$ & $58.9$ & $61.1$ & $54.2$ & \textbf{89.9} & 88.9\\
Page-header & $\operatorname{85-89}$ & $70.0$ & $72.0$ & $67.9$ & $63.7$ & 70.4 & \textbf{90.5}\\
Picture & $\operatorname{69-71}$ & $72.7$ & $72.0$ & $77.1$ & $\textbf{84.1}$ & 82.0 & 82.4\\
Section-header & $\operatorname{83-84}$ & $69.3$ & $68.4$ & $74.6$ & $64.3$ & 86.2 & \textbf{87.7}\\
Table & $\operatorname{77-81}$ & $82.9$ & $82.2$ & $\textbf{86.3}$ & $85.7$ & 84.4 & 85.3\\
Text & $\operatorname{84-86}$ & $85.8$ & $85.4$ & $\textbf{88.1}$ & $83.3$ & 86.1 & 83.5\\
Title & $\operatorname{60-72}$ & $80.4$ & $79.9$ & $82.7$ & 82.8 & 75.3 & \textbf{83.4}\\
\hline mAP & $\operatorname{82-83}$ & $73.5$ & $73.4$ & $76.8$ & $74.3$ & 79.6 & \textbf{83.8}\\
\hline
\end{tabular}
\end{center}

\label{tab:doclaynet-pod}
\end{table}

\subsubsection{DocLayNet.} We benchmark our approach against other highly competitive vision-based methods, including state-of-the-art object detection methods such as DINO \cite{zhang2022dino} and YOLOv5 \cite{yolov5}, as well as the hybrid approach of vision-only DOC \cite{wang2023doc}. As shown in Table~\ref{tab:doclaynet-pod}, our approach surpasses all those top-down based methods. Furthermore, in comparison with DOC, \Ours{} notably enhances the mAP from 79.6\% to 83.8\%.
\Ours{} notably excels in several categories, such as formula and page-header, demonstrating significant performance improvements over prior methods. This emphasizes its effectiveness in seamlessly incorporating the features of relevant text-lines for graphical object detection, while also adeptly capturing broader layout information to improve logical role classification. 
Given that DocLayNet is an exceptionally challenging dataset, featuring diverse document scenarios and numerous text regions with fine-grained logical roles, the superior performance of \Ours{} highlights its advantage in handling complex document layouts.
While \Ours{} generally outperforms DOC, it shows lower performance in certain categories like Footnote. This is mainly due to multi-stage methods, like DOC, leveraging predicted graphical object boxes in advance to filter text-lines within graphical objects, effectively addressing the long-tail problem. Due to the visual pattern similarities between text lines within graphical objects and predefined standard text lines, current classification methods can inadvertently impact the performance of predefined standard text categories.
In the future, we plan to enhance \Ours{} by refining the training strategy and incorporating additional relationship definitions to mitigate this issue.

\subsection{Ablation Studies}
\begin{table}[t]  


\centering
\caption{Ablation study of Type-wise Queries in page object detection task on Comp-HRDoc (in \%).}
\begin{adjustbox}{width=\textwidth}
\begin{tabular}{
    l|cccccccccccc|c
}
\toprule
Method & {Title} & {Author} & {Mail} & {Affil} & {Sect} & {Para} & {Table} & {Fig} & {Cap} & {Foot} & {Head} & {Footn} & {Segm. mAP} \\
\midrule
DLAFormer & 96.8 & 91.1 & 75.4 & 86.0 & 92.8 & 87.7 & 79.4 & 85.5 & 92.5 & 97.8 & 98.2 & 91.7 & \textbf{89.6} \\
\quad - Type-wise Queries & 96.0 & 90.7 & 67.8 & 83.8 & 91.5 & 87.1 & 77.4 & 84.7 & 92.3 & 97.8 & 98.2 & 91.5 & 88.2 \\ 
\bottomrule
\end{tabular}
\end{adjustbox}
\label{tab:abl-type-1}
\end{table}

\begin{table}[t]  

\fontsize{9}{11}\selectfont
\setlength{\tabcolsep}{4pt}

\centering
\caption{Ablation study of Type-wise Queries in reading order prediction task on Comp-HRDoc (in \%).}
\begin{tabular}{lcc}
\toprule
{ Method}         & { \begin{tabular}[c]{@{}c@{}}Text Region \\ Edit Distance Score\end{tabular}} & { \begin{tabular}[c]{@{}c@{}}Graphical Region \\ Edit Distance Score\end{tabular}} \\ \midrule
{ DLAFormer-R50}       & { \textbf{96.6}}                                                                      & { \textbf{90.0}}                                                                          \\ 
{ \quad - Type-wise Queries}             & { 96.6}                                                                              & { 89.4}                                                                                  \\
\bottomrule
\end{tabular}

\label{tab:abl-type-2}
\end{table}
\subsubsection{Effectiveness of Type-wise Queries.} To validate our proposed type-wise queries, we conduct experiments replacing them with learnable content queries, akin to DINO \cite{zhang2022dino}. As shown in Table~\ref{tab:abl-type-1}, type-wise queries prove beneficial not only for tables and figures regression but also for logical role classification in text regions. This implies that introducing type-wise queries as content queries effectively enhances their physical meaning, facilitating adaptive feature capture from distinct regions for various query types. In the reading order prediction task, removing type-wise queries from \Ours{} results in a 0.7\% decrease in Graphical REDS on Comp-HRDoc, as seen in Table~\ref{tab:abl-type-2}. This indicates that leveraging type-wise queries as content queries helps explore logical relationships between different region types.

\begin{table}[t]

\fontsize{9}{11}\selectfont

\centering  
\caption{Ablation study of transformer architectures in \Ours{} on Comp-HRDoc (in \%).}
\begin{adjustbox}{width=\textwidth}
\begin{tabular}{c|ccc@{\hspace{4pt}}|ccc}
\toprule
\# &  Encoder & Decoder & Feature Map Scales & \begin{tabular}[c]{@{}c@{}}Segm. \\ mAP\end{tabular}              & \begin{tabular}[c]{@{}c@{}}Text Region \\ REDS\end{tabular} & \begin{tabular}[c]{@{}c@{}}Graphical Region\\ REDS\end{tabular} \\ \midrule
1 & DETR         & DAB-DETR                 & 1/4                                 & 86.9                  & 96.2                                                                     & 88.9                                                                        \\
2 & DETR         & Deformable-DETR          & 1/8,1/16,1/32,1/64                  & 89.0                  & 96.4                                                                     & 89.4                                                                        \\
3 & Deformable-DETR          & Deformable-DETR          & 1/8,1/16,1/32,1/64                 & \textbf{89.6}                  & \textbf{96.6}                                                                     & \textbf{90.0} \\ \bottomrule                                                                       
\end{tabular}
\end{adjustbox}
\label{tab:abl-arch}
\end{table}



\subsubsection{Effectiveness of Deformable DETR based Architecture.} As summarized in Table~\ref{tab:abl-arch}, we perform ablation experiments to investigate the transformer architecture in \Ours{}. Leveraging efficient support for multi-scale features and simultaneous consideration of global and local information in deformable cross-attention, \Ours{} with deformable DETR \cite{zhu2020deformable} decoder outperforms its counterpart with DAB-DETR \cite{liu2022dab} decoder across both page object detection and reading order prediction tasks, as presented in rows 1 and 2 of Table~\ref{tab:abl-arch}. Additionally, replacing the vanilla DETR encoder with a deformable encoder further enhances the performance of \Ours{}, as depicted in rows 2 and 3.

\section{Conclusion and Future Work}

In this paper, we introduce \Ours{}, a novel end-to-end transformer-based approach for document layout analysis, consolidating tasks including graphical page object detection, text region detection, logical role classification, and reading order prediction within a unified model. To achieve this, we treat various DLA sub-tasks as relation prediction problems and propose a unified label space approach to enabling a unified relation prediction module to effectively and efficiently handle these tasks concurrently. Comprehensive experiments underscore the superiority of our unified model over existing methodologies that employ multi-branch or multi-stage structures to address these DLA tasks. 
Our results indicate that label unification and relation prediction are promising avenues for future research in the DLA field.

In future work, we plan to evolve the unified model to encompass additional document-level DLA tasks, including table of contents extraction and hierarchical document structure reconstruction. We also aim to expand the unified label space approach for a wider array of tasks and integrate text embedding to augment the model's capabilities. These advancements are anticipated to significantly bolster the model's effectiveness in document layout analysis. 
Furthermore, given \Ours{}'s proficiency in simultaneous localization and relationship prediction, its application in a variety of tasks fitting this paradigm, such as visual relationship detection, is an area meriting in-depth investigation.
%
%
%
\bibliographystyle{splncs04}
\bibliography{bibfile}
\clearpage
\end{document}